\documentclass [review]{elsarticle} 

\usepackage{lineno,hyperref}
\modulolinenumbers[5]
\usepackage{float}
\usepackage{graphicx}
\usepackage{subfig}
\usepackage{mathrsfs}
\usepackage{graphics}
\usepackage{float}
\usepackage{amsmath}
\usepackage{amsfonts}
\usepackage{optidef}
\usepackage[subnum]{cases}

\usepackage{xcolor}

\usepackage{xcolor}
\hypersetup{
    colorlinks,
    linkcolor={red!50!red},
    citecolor={blue!50!blue},
    urlcolor={blue!80!blue}
}

\usepackage{optidef}
\usepackage{graphicx}
\usepackage{float}
\usepackage{graphicx}
\usepackage{subfig}
\usepackage{mathrsfs}
\usepackage{float}
\usepackage{amsmath}
\usepackage{amsfonts}
\usepackage{booktabs}
\usepackage{algorithm}
\usepackage[noend]{algpseudocode}
\usepackage{multirow}
\algdef{SE}[DOWHILE]{Do}{doWhile}{\algorithmicdo}[1]{\algorithmicwhile\ #1}

\usepackage[letterpaper, portrait, margin=1in]{geometry}

\usepackage{optidef}

\usepackage{amsthm}

\theoremstyle{definition}

\makeatletter
\newcommand*{\centerfloat}{%
  \parindent \z@
  \leftskip \z@ \@plus 1fil \@minus \marginparwidth
  \rightskip \leftskip
  \parfillskip \z@skip}
\makeatother

\usepackage{setspace}
\usepackage{natbib}
\usepackage{algorithm}
\usepackage[noend]{algpseudocode}
\algdef{SE}[DOWHILE]{Do}{doWhile}{\algorithmicdo}[1]{\algorithmicwhile\ #1}

\usepackage{appendix}

\usepackage{array}
\usepackage{threeparttable}
\usepackage{multirow}
\usepackage{rotating}
\usepackage{makecell}

\usepackage{subfig}

\usepackage{url}

\newsavebox{\measurebox}

\journal{}

\biboptions{authoryear}

\begin{document}
\begin{frontmatter}

\title{Efficient Textual Explanations for Complex Road and Traffic Scenarios \\based on Semantic Segmentation}


\author[label1,label2]{Yiyue Zhao}\corref{coauthornote}\ead{1952430@tongji.edu.cn}
\author[label3]{Xinyu Yun}\corref{coauthornote}\ead{1953217@tongji.edu.cn}
\author[label1,label2]{Chen Chai\corref{mycorrespondingauthor}}\ead{chaichen@tongji.edu.cn}
\author[label3]{Zhiyu Liu}\corref{coauthornote}\ead{1952668@tongji.edu.cn}
\author[label1,label2]{Wenxuan Fan}\corref{coauthornote}\ead{1951508@tongji.edu.cn}
\author[label1,label2]{Xiao Luo}\corref{coauthornote}\ead{luo.xiao@tongji.edu.cn}
\address[label1]{{College of Transportation Engineering, Tongji University, Shanghai, 201804, China}}
\address[label2]{{The Key Laboratory of Road and Traffic Engineering of the Ministry of Education, Tongji University, Shanghai, 201804, China}}
\address[label3]{{School of Economics and Management, Tongji University, Shanghai, 200092, China}}
\cortext[mycorrespondingauthor]{Corresponding author}


\begin{abstract}

The complex driving environment brings great challenges to the visual perception of autonomous vehicles. It's essential to extract clear and explainable information from the complex road and traffic scenarios and offer clues to decision and control. However, the previous scene explanation had been implemented as a separate model. The black box model makes it difficult to interpret the driving environment. It cannot detect comprehensive textual information and requires a high computational load and time consumption. Thus, this study proposed a comprehensive and efficient textual explanation model. From 336k video frames of the driving environment, critical images of complex road and traffic scenarios were selected into a dataset. Through transfer learning, this study established an accurate and efficient segmentation model to obtain the critical traffic elements in the environment. Based on the XGBoost algorithm, a comprehensive model was developed. The model provided textual information about states of traffic elements, the motion of conflict objects, and scenario complexity. The approach was verified on the real-world road. It improved the perception accuracy of critical traffic elements to 78.8\%. The time consumption reached 13 minutes for each epoch, which was 11.5 times more efficient than the pre-trained network. The textual information analyzed from the model was also accordant with reality. The findings offer clear and explainable information about the complex driving environment, which lays a foundation for subsequent decision and control. It can improve the visual perception ability and enrich the prior knowledge and judgments of complex traffic situations. 

\end{abstract}



\end{frontmatter}


\section{Introduction}\label{intro}

Accurate perception and textual explanation for the road environment lay a solid foundation for the development of autonomous vehicles(AVs). Multi-source sensors have been applied to scene perception, including LIDAR, RADAR, visual sensors, etc. As visual perception has many advantages including the richness of features, low cost, and higher resolution, it has been applied in AVs widely\citep{understanding}. Influenced by adverse weather conditions and uncertain traffic flow, the accuracy and efficiency of visual perception drop off sharply\citep{behavior}. And the perception failure causality is still unclear. It's essential to interpret what AVs detect.
With the rapid development of deep learning networks, end-to-end and pipelined systems have emerged to explain the driving environment. The former transforms the visual input into the object's action directly while the latter decomposes the perception problem into several sub-problems such as detecting, tracking, and planning\citep{2020explainable}. However, both of them require large datasets and complex frameworks, which decreases the efficiency of visual perception\citep{2019multi}. 

But many researchers regarded scene explanation as a semantic segmentation problem\citep{efficient}. Previous studies applied various deep learning networks to assign semantic labels to individual segments and detect the edge and semantic information. It could distinguish objects of different classes. However, segmentation just detects the static objects in the road environment, but they cannot explain more dynamic information like potential hazard awareness and complexity evaluation\citep{explanations}. Because of the simple training sample\citep{WOS:000432492800006}, segmentation networks have difficulty detecting the obstructed and small objects in the complex road and traffic scenarios. Besides, the endless changes in the road environment make it difficult for visual perception to be applied on a large scale.

Thus, this paper proposed a comprehensive model based on semantic segmentation to explain the textual information of complex road and traffic scenarios. By using vehicle-mounted sensors of Tesla, this study collected images of complex road and traffic scenarios and established a dataset. To obtain clear class information efficiently, this study constructed a semantic segmentation algorithm to extract  traffic elements and analyzed their textual states. Furthermore, XGBoost algorithm was applied to classify the traffic scenarios such as car-following, lane change, emergency avoidance, etc. Based on temporal difference algorithm, the motion of conflict objects was estimated, which provided accurate clues for decision and control. Finally, this study evaluated scenario complexity to quantify safety level dynamically. The whole approach was also verified on the real-world road. 

The major contributions of this paper include:

(1) This study proposes a method to explain textual information from the perspective of visual perception. It provides clear and explainable situation awareness of complex road and traffic scenarios.

(2) Through the transfer learning method, this study improves the accuracy and efficiency of semantic segmentation in complex road and traffic scenarios. The lightweight and efficient structures allow the method to be flexibly adapted to various environments.

The rest of this paper is organized as follows. Section 2 summarizes related work about road semantic segmentation and textual explanation methods. Section 3 covers the critical methodology of textual explanation. Section 4 verifies the approach on the real-world road. Finally, Section 5 concludes this paper with contributions, discussions, and future studies.

\section{Related Works}

\subsection{Road semantic segmentation algorithms}

In the early ages of semantic segmentation research, it was mainly used to detect body joints’ 3D location based on TextonForest\citep{4587503}, Random Forest Classifier\citep{Ellis2014A}, and other traditional machine learning methods. With the rapid development of artificial intelligence, the deep learning framework was introduced into semantic segmentation research to enhance accuracy and synchronicity.
 
Due to the advantages of low cost and global detection, semantic segmentation has been applied widely to visual perception and textual explanation. It assigns each pixel of the road scenario image with a class label. So road scenario could be transformed into pixel-level semantic information and recognized by a visual perception system\citep{SULTANA2020106062}. State-of-the-art segmentation algorithms could gain multi-layer contextual information about the surrounding environment, which provides accurate data for the proper planning and controlling of autonomous vehicles\citep{2021Review}. 


To solve the contradiction between the expedition of the convolutional kernel sensory field and the missing location information, DeconvNnet\citep{7410535} and Dilated Convolutions\citep{yu2016multiscale} were introduced in 2015. These algorithms created convolution-deconvolution layers and dilated convolution networks, which could operate multi-dimensional space information and enhance the ability to extract location information in the margin. In 2016, DeepLab v2\citep{2018DeepLab} and PSPNet\citep{8100143} were created to better obtain global context information with local features in the road scenarios, which made semantic information and edge information of road scenarios more subsequent\citep{10.1007/978-3-030-01234-2_49}. And algorithms have gradually developed into weakly supervised and unsupervised fields\citep{2021Review}.

Although researchers have been optimizing the segmentation network structures and parameters, the accuracy and efficiency still need to be improved. The cost of every trial is huge since Convolutional Neural Network(CNN) requires a large amount of calculation power and time. For example, it took five days for Long et al.\citep{2015Fully} to refine the FCN-8s model on a single GPU (NVIDIA Tesla K40). 


\subsection{Datasets of road semantic segmentation}


Many researchers have utilized multi-source sensors to capture multi-dimensional information on the real-world road and constructed large datasets of urban road scenes like Cityscapes, KITTI, CamVid, etc. 

Cityscapes, a prominent dataset that focuses on semantic understanding of urban street scenes, contains various stereo video sequences recorded from street scenarios from 50 different cities. It consists of high-quality pixel-level annotations of 5000 frames and a larger set of 20,000 weakly annotated frames. Thus, the dataset provides annotations with more details than other open-source datasets. However, it contains only simple urban street scenes that lack comprehensively annotated road scene information\citep{WOS:000432492800006}. And the very simple samples make it easy to arrive at the bottleneck\citep{Multi-Label}. 

Due to the uncertain traffic participants and endless changes of elements, road and traffic scenarios have become more complex. The complexity can be attributed to two aspects:
 
(1) Structures of the complex road environment such as the large flyover, overhead, circle island, and crossing bridge. They are quite different from common ground road so the visual perception is likely to identify them to ground road and cause collision accidents. Complex ground signs and signal phases also increase the perception difficulty.


(2) Uncertain and dense pedestrian and non-motorized vehicles flow. Their behaviors and trajectories are unpredictable. Besides, pedestrian and non-motorized vehicles are often hidden by surrounding infrastructures, bringing a great challenge to visual perception.


\subsection{Textual explanation methods}

Existing textual explanation methods majorly depend on a separate approach of visual perception, including semantic segmentation, object detection, trajectory analysis, etc.\citep{2020explainable}\citep{efficient}\citep{Discriminative} However, these approaches pay more attention to high-level contextual elements such as sky, walls, and vegetation\citep{closing}. These elements are not vital to the safety of complex road and traffic scenarios\citep{efficient}. 

To obtain more useful textual information of the road environments, scenario characterization becomes an important issue\citep{understanding}. Scenario complexity evaluation, motion estimation of conflict objects, and road type analysis are the critical measures\citep{WOS:000656963900005}.

Scenario complexity can be evaluated by three aspects: the quantity, variety, and relation of elements\citep{WOS:000656963900005,towards}. Quantity mainly includes the number of static elements. The variety infers diverse traffic infrastructures or dynamic features while the relation means the interconnection or conflict among traffic constitution.
The static quantity is fixed in the road scenario and it’s easy to calculate the number of elements by visual perception. But the dynamic feature is hard to measure. The existing models evaluate the collision risk and safety field based on vehicles’ trajectory data\citep{WOS:000617351800001}. It requires plenty of V2V communication and data processing\citep{WOS:000719424500241}, so the efficiency and lighter data have become the bottleneck.







\section{Methodology} \label{method}

To establish an efficient textual explanation approach, this paper first collected the data on the real-world road. Then images of complex road and traffic scenarios were extracted and labeled to construct a dataset. Next, this study constructed a semantic segmentation model and improved the accuracy and efficiency by the transfer learning method. Based on the segmentation results and XGBoost, this study developed a comprehensive textual explanation model for complex road and traffic scenarios.

\subsection{Data acquisition} \label{network_repre}
To make the segmentation algorithm better adapt to the complex road and traffic scenarios, this study constructed a dataset. The process of data acquisition could be divided into three steps:

Step1: Collect data on the real-world road. Through a monocular sensor of Tesla installed on the hood, videos of the driving process could be recorded. According to the requirement of visual perception, this study selected videos in the front direction.

Step2: Extract images of complex road scenarios. To increase the complexity of the datasets, this study extracted specific complex scenarios. Based on the definition of complex road scenarios in Section 2.2, this study checked the collected videos frame by frame and chose 271 images including complex scenarios. 

Step3: Assign each element with a class label in the selected images. Based on Labelme(a third-party package of Python), this study sketched the contour of objects manually and assign them with corresponding labels. Then JSON files were transformed into colorful images by assigning each class with a specific color.

Step4: image gray-scaling. Colorful images occupy a large amount of space to store the RGB values, which declines the training efficiency. So this paper transformed them into gray-scale images by Equation \ref{eq:gray}:

\begin{equation}
G=\sqrt{\frac{R^{2.2}+(1.5G)^{2.2}+(0.6B)^{2.2}}{1+1.5^{2.2}+0.6^{2.2}}}
\label{eq:gray}
\end{equation}

Then, this study reassigned a gray-scale value to each class from 0 to 255 by a gradient. It could distinguish different classes in the training process. And semantic information of each pixel was also transformed to gray-scale value simultaneously.

\subsection{Road semantic segmentation algorithm}

Accurate semantic segmentation understands the boundary information of the road environment. It offers class information of traffic elements in the surrounding driving environment. So this study established a semantic segmentation algorithm and improved its accuracy and efficiency based on transfer learning. 

\subsubsection{Global network}

Pyramid Scene Parsing Network(PSPNet) was selected as the global backbone. The segmentation network applied spatial pyramid pooling to obtain statistics on spatial scenes. It provided a good description for the interpretation of the overall road scenarios\citep{7780459}. In detail, the input layers were divided into 6×6, 3×3, 2×2, and 1×1 grids\citep{zhou2015object}. The features obtained from each layer were integrated and subsequently executed in one convolution operation to obtain the predicted image\citep{2017arXiv170404861H}. 

\begin{figure}[H]
\centering
\includegraphics[width= 1 \textwidth]{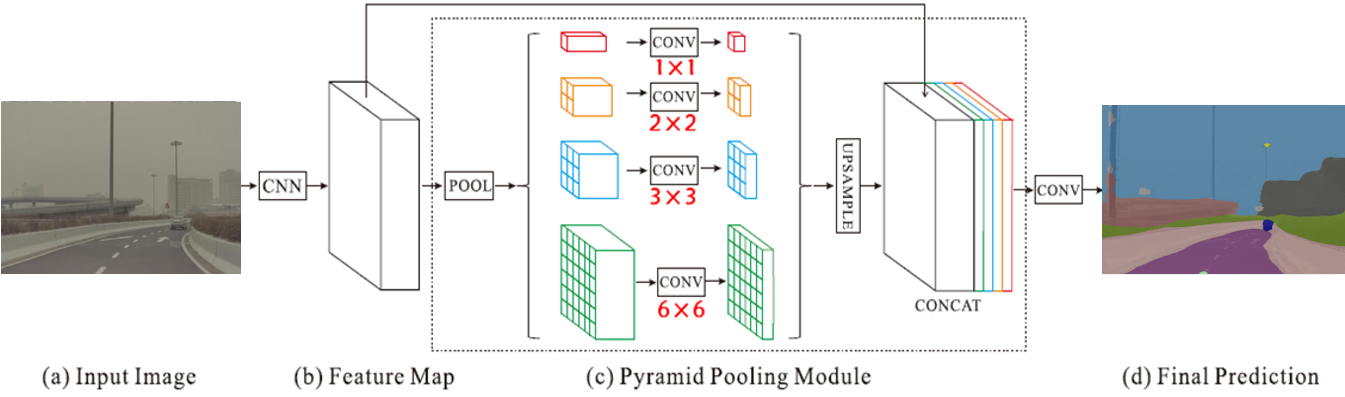}
\caption{Global network of the road semantic segmentation algorithm.}
\label{fig:psp}
\end{figure}

And this study added a ResNet50 structure into the feature enhancement extraction network. It condensed the image’s dimension before the convolutional layer and expanded the dimension after a convolution operation. The optimized structure-bottleneck design could perceive smaller objects and capture the edge information more accurately. Figure \ref{fig:psp} shows the global network of the road semantic segmentation algorithm. 

\subsubsection{Evaluation metrics}
Road semantic segmentation is typically a classification problem. The probability distribution of classes is usually the output in classification models. This paper applied the cross-entropy loss function to evaluate the deviation between the true distribution and the predicted ones by the segmentation model \citep{WOS:000722273400048}. The smaller the loss value is, the more accurate the model prediction is. The discrete form of the cross-entropy function is shown as Equation \ref{eq:entro1}.

\begin{equation}
L(y,f(x))=H(p,q)=-\frac{1}{N}\sum_{i=1}^{N}p(y_i|x_i)log[q(\hat{y_i}|x_i)]
\label{eq:entro1}
\end{equation}

where 

\qquad $p$ is the actual distribution of data label

\qquad $q$ is the predicted distribution of data label

\qquad $p(y_i|x_i)$ is the actual distribution of sample $x_i$

\qquad $q(y_i|x_i)$ is the probability distribution of sample $x_i$ on each class

In general, sample $x_i$ only belongs to a certain class $c_k$, so $p(\hat{y_i}=c_k|x_i)=1$. It means that the probability of the other classes is zero. Then the above equation could be simplified as Equation \ref{eq:entro2}:

\begin{equation}
L(y,f(x))=H(p,q)=-\frac{1}{N}\sum_{i=1}^{N}p(y_i|x_i)log[q(\hat{y_i}=c_k|x_i)]
\label{eq:entro2}
\end{equation}

As the category was determined as $c_k$,  minimizing the loss value of Equation \ref{eq:entro2} was equivalent to finding the maximum likelihood probability value of the label.

Based on the cross-entropy value of each pixel, the general perception accuracy of an image was obtained. In this paper, mean intersection and merging ratio($mIoU$) was selected as the quantitative evaluation index. It evaluates the accuracy of each class from a global perspective. The $mIoU$ is calculated by Equation \ref{eq:miou}:

\begin{equation}
mIoU=\frac{1}{k+1}\sum_{i=0}^{k}\frac{p_{ii}}{\sum_{j=0}^{k}p_{ij}+\sum_{j=0}^{k}p_{ji}-p{ii}}
\label{eq:miou}
\end{equation}

where

\qquad $k$ is the number of classes

\qquad $p_{ii}$ is the target number for which both the predicted and true values are $i$

\qquad $p_{ij}$ is the number for which the predicted value is $j$ and the true value is $i$

\qquad $p_{ji}$ is the number for which the predicted value is $i$ and the true value is $j$

\subsubsection{Network improvement based on transfer learning} \label{sec:trans}

To better enhance the accuracy and efficiency of road semantic segmentation, this paper applied the transfer learning method to improve the initial network structure. 

Transfer learning migrates the trained parameters from the source domain to the target domain\citep{2021Transfer}. It utilizes the learning results of the pre-trained model effectively and requires a smaller volume of dataset\citep{2019fine}. So the improved model can rapidly improve the training efficiency. 

In this paper, the transfer learning method could be divided into two steps:

Step1: Transfer data from the open-source dataset Cityscapes to the complex road dataset. Because the pre-trained model was trained based on simple scenario samples, it cannot adapt to perceive some elements in complex road scenarios perfectly. Thus, this paper utilized the transfer learning method and established the association between the two datasets, including the "many-to-one" mapping relationship and the "one-to-one" derivation relationship. Through data transfer, this study removed the extra elements (not related to traffic safety) from the driving environment. Then the perception of critical traffic elements will be more specific.

Step2: Optimize the hyper-parameters of the pre-trained model. It's difficult for the pre-trained model to recognize small-size conflict objects (like pedestrians and non-motorized vehicles) in a complex scenario. So this paper optimized the hyper-parameters in the pre-trained model to improve training efficiency and accuracy. The optimized hyper-parameters are shown below:

\begin{itemize}
\item Input size: expanding to 2048$\times$1024, to improve boundary segmentation accuracy.

\item Learning rate: applying cosine anneal to find the global optimal solution rapidly and accurately.

\item Batch size and the number of workers: increasing them to enhance training efficiency.

\item Regular term: adding the L1 and L2 regular terms to prevent overfitting.

\begin{equation}
\Omega(f)=\gamma T+\frac{1}{2}\lambda \| \omega \|^2
\label{eq:regular}
\end{equation}

$\Omega(f)$ limits the number of leaf nodes of the tree to prevent the classification decision tree from growing too large. The smaller the $\Omega(f)$ value is, the stronger the generalization ability model has.

\item Momentum term: adding momentum to prevent local minimal error.

\begin{equation}
\Delta_{p}w_{ij}(t+1)=\alpha \Delta_{p} w_{ij}(t)+\eta \delta_{pj}+o_{pj}
\label{eq:moment}
\end{equation}

where

\qquad $\eta$: learning step coefficient, this paper chooses 0.1

\qquad $\alpha$: stable coefficient, this paper chooses 0.8

Based on the back-propagation method, this study added a value $\eta \delta_{pj}+o_{pj}$  proportional to the last weight change to each weight change, which was used to slip through the local minimal error.

\end{itemize}

\subsection{Textual explanation algorithm}

\subsubsection{Scenario classification}
\begin{figure}[H]
\centering
\includegraphics[width= 0.97 \textwidth]{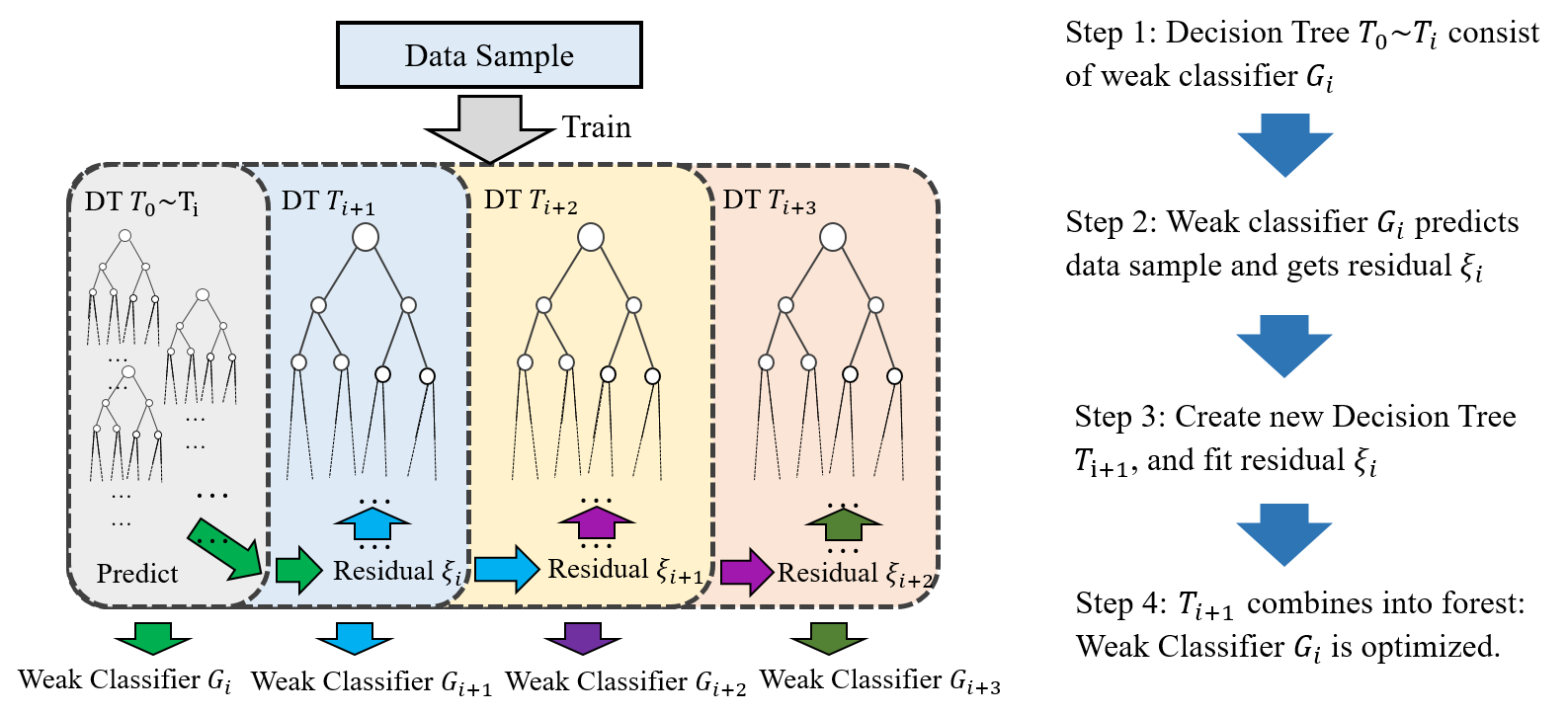}
\caption{Framework of ensemble learning method: XGBoost.}
\label{fig:xgb}
\end{figure}

This study applied the XGBoost algorithm \citep{2016} to construct a scenario classification decision tree, as seen in Figure \ref{fig:xgb}. The detailed algorithm was shown as follows.

(1)Decompose elements in complex scenarios and add new decision trees into the forest. It makes the model learn a new function $f(x)$ to fit the residual of the last decision tree.

\begin{equation}
\begin{aligned}
\hat{y}&=\phi(x_i)=\sum_{k=1}^{K}f_k(x_i) \\
where \quad F&={f(x)=\omega_{q(x)}},q:R^m \rightarrow T,\omega \in R
\end{aligned}
\label{eq:residual}
\end{equation}

(2) After establishing $K$ regression trees, this study obtained the predicted score of elements in scenarios from the root node to leaf node $q$.

(3) Summing up the predicted scores of regression trees and assigning images with scenario labels. 

\subsubsection{Conflict object's motion detection}

In the complex driving environment, kinematic state detection of conflict objects offers the safety level dynamically. Based on processed segmentation results in Figure \ref{fig:data}, conflict objects’ location distribution could be obtained. Choosing the DBSCAN algorithm to identify the clustering center $(x_i,y_i)$, this paper regarded the center trajectory as the objects’ movement trajectory. Through temporal differences between two adjacent frames, this study gained the relative speed in longitudinal and lateral directions:

\begin{equation}
\begin{aligned}
v_{x_{i+1}}=fps \cdot (x_{i+1}-x_i)\quad,\quad v_{y_{i+1}}&=fps \cdot (y_{i+1}-y_i) \\
a_{x_{i+1}}=fps^2 \cdot (x_{i+2}+x_i-2x_{i+1})\quad,\quad a_{y_{i+1}}&=fps^2 \cdot (y_{i+2}+y_i-2y_{i+1})
\end{aligned}
\label{eq:state}
\end{equation}

Furthermore, this paper calculated time to collision (TTC) to measure the urgent degree of the conflict intersection\citep{ttc}. The first order of TTC was calculated by Equation \ref{eq:ttc}:

\begin{equation}
TTC=\frac{y_i}{v_{y_i}}
\label{eq:ttc}
\end{equation}

Muller found that the severe conflict threshold corresponded to 1.0 second of TTC\citep{ttc}. So this study added $\frac{1}{TTC}$ to the relation complexity. The more TTC is, the less urgent the conflict is.

\subsubsection{Scenario complexity evaluation}

In general, XGBoost provided the probability distribution of each label. So this paper defined the relation complexity through the expectation of scenario probability:

\begin{equation}
C=E(s(i))=\sum_{i=1}^{4}s(i)p(i)
\label{eq:complexity}
\end{equation}

where

\qquad $s(i)$ is the relation complexity of the scenario

\qquad $p(i)$ is the probability of scenario label

According to the accident risk rate analyzed by Shanghai Naturalistic Datasets(SH-NDS)\citep{2019Analysis}, the value of $s(i)$ was defined in Table \ref{tab:label}(min=1, max=5):

\begin{table}[H]
\centering
\caption{The definitions and values of the relation complexity}
\begin{tabular}{@{}ccccc@{}}
\toprule
Scenario &	Free driving &	Following &	Cutting in &	Emergency avoidance \\
\midrule
$s(i)$	& 1	&  3  &	 4  & 	5 \\
\bottomrule
\end{tabular}
\label{tab:label}
\end{table}

According to the segmentation result, the most number of classes among the self-built dataset was defined as the quantity complexity, i.e. $n_{max}$. And variety complexity could be defined as segmentation accuracy $m\%$. The more accurate the model segments, the lower the variety complexity is. From the perspectives of quantity, variety, and relation complexity, this paper calculated the scenario complexity $d$:

\begin{equation}
d=C\cdot [(1-\frac{m}{100})+\frac{n}{n_{max}}+\frac{1}{TTC}]
\label{eq:complex}
\end{equation}


\section{Validation}

Based on the methodology in Section 3, this paper validated the effect of the textual explanation approach on the real-world road in China.

\subsection{Semantic segmentation of complex road and traffic scenarios}

\subsubsection{Setup}

\begin{figure}[H]
\centering
\subfloat[Adverse weather conditions]{\includegraphics[width=0.3\textwidth]{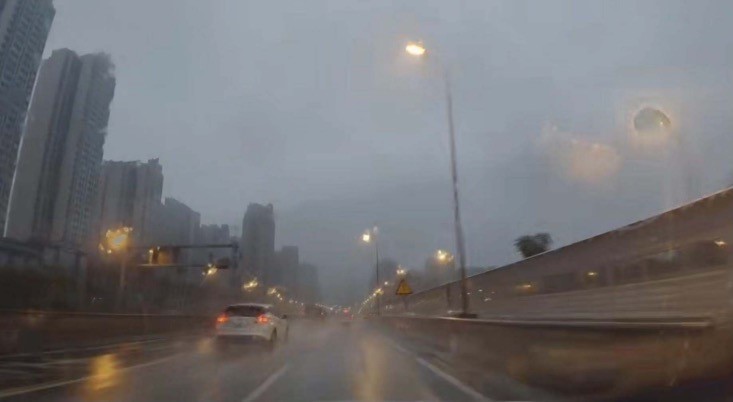}\label{weather}}
\hfil
\subfloat[Complex road infrastructures]{\includegraphics[width=0.327\textwidth]{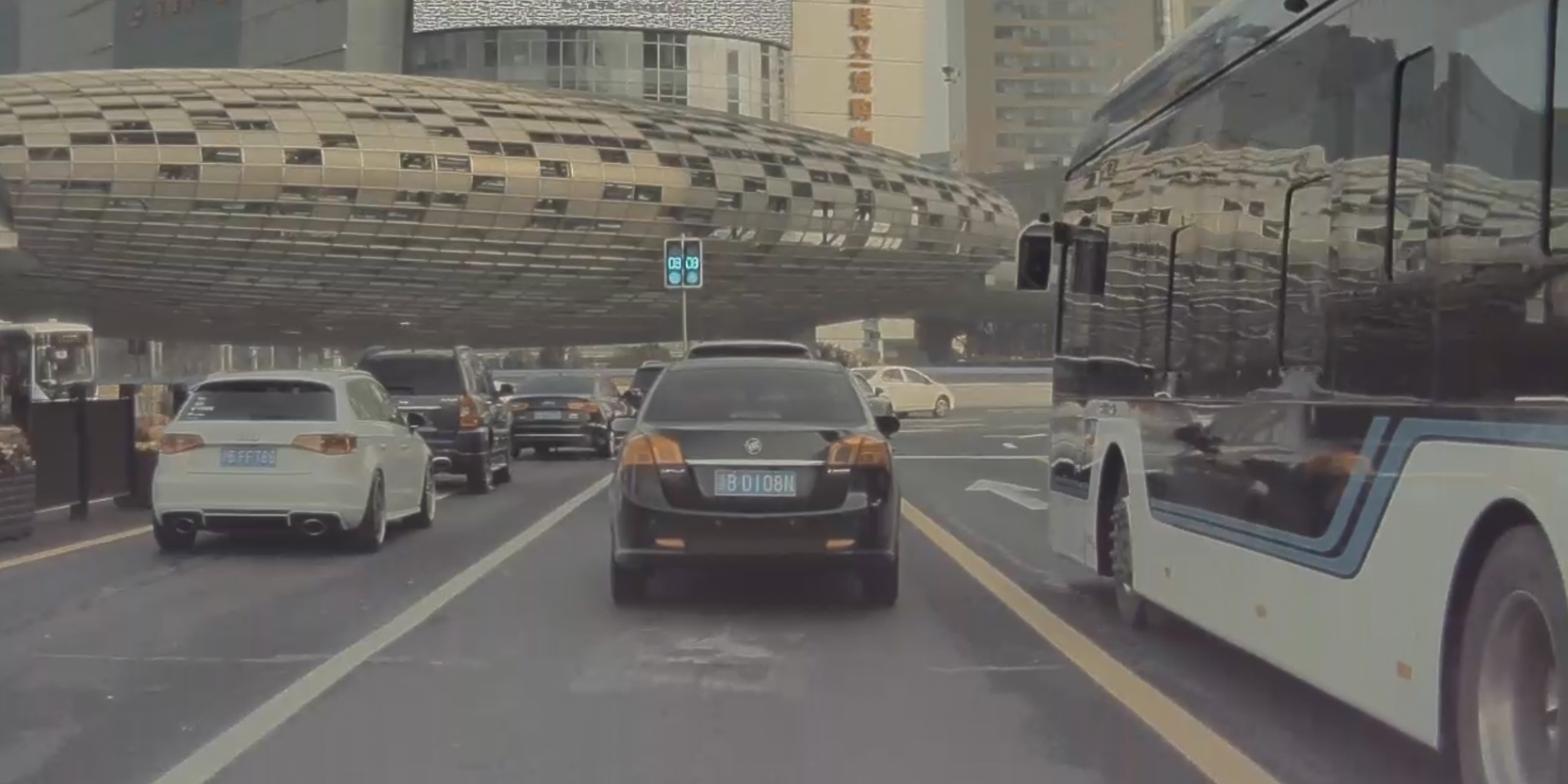}\label{infra}}
\hfil
\subfloat[Mixed traffic flow]{\includegraphics[width=0.3\textwidth]{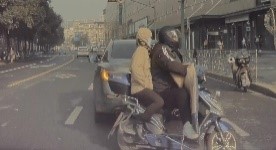}\label{intersection}}
\caption{The example of images in complex road and traffic scenarios.}
\label{fig:dataset}
\end{figure}

This study used a Tesla vehicle-mounted monocular sensor to capture videos in Shanghai, China. 640 video sequences covered various traffic scenarios such as ramp merging, cut-in, and intersection conflict. The size of the video sequences was over 8GB, containing 336k frames of images. By comparing open-source datasets with the road environmental portraits of 34 cities in China, the typical frames of the complex driving environment were extracted. They included complex road infrastructures, dense non-motorized vehicle flow, and random pedestrian flow. For a series of continuous frames, a critical frame can represent road and traffic scenarios at that moment because of few spatio-temporal changes. Then a total of 271 images were selected into a dataset. By transfer learning, the semantic segmentation network could inherit features of complex driving environments rapidly based on a small volume of training data\citep{2018retinal}. The examples of the dataset are shown in Figure \ref{fig:dataset}.

Then the objects of images were assigned with class labels by Labelme. According to the traffic attribute of elements, objects were assigned to 22 classes. Table \ref{tab:class} shows the examples of classification definitions. 

\begin{table}[H]
\centering
\caption{Class and corresponding definitions of the complex scenario dataset}
\begin{tabular}{@{}ccc@{}}
\toprule
Class	&	Class ID	&	Definition	\\
\midrule
Background	&	0	&	Void, ground	\\
Barrier	&	1	&	Obstacles between vehicle and pedestrian	\\
Sign	&	5	&	Traffic signals	\\
Car	&	7	&	All kinds of vehicles	\\
nmt	&	8	&	Non-motorized vehicles	\\
infra	&	10	&	Road infrastructure	\\
Road line	&	12	&	Road line signal	\\
Tree	&	15	&	Woods and bushes	\\
road	&	18	&	Traffic road and pedestrian road	\\
pole	&	22	&	Light pole	\\
\bottomrule
\end{tabular}
\label{tab:class}
\end{table}

To achieve the normalization requirement of the input data in the convolutional neural network, this paper processed the images in the dataset: 

\begin{itemize}

\item Adjusting the input size of images to 1024 $\times$ 512.

\item Transforming the colorful images to the gray-scale images by Equation \ref{eq:gray}. 
\end{itemize}

\begin{figure}[H]
\centering
\subfloat[Initial image]{\includegraphics[width=0.3\textwidth]{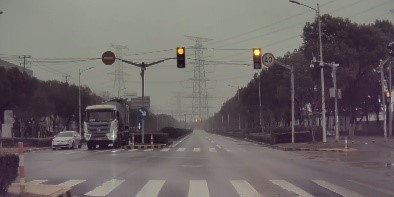}\label{initial}}
\hfil
\subfloat[Labeled image]{\includegraphics[width=0.3\textwidth]{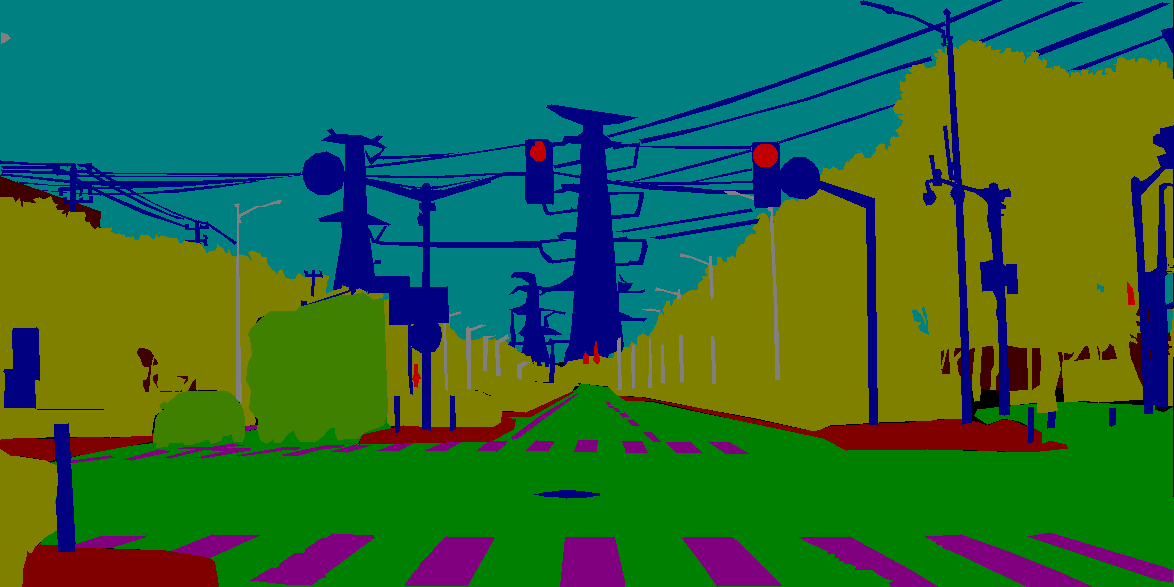}\label{labeled}}
\hfil
\subfloat[Gray-scale image]{\includegraphics[width=0.3\textwidth]{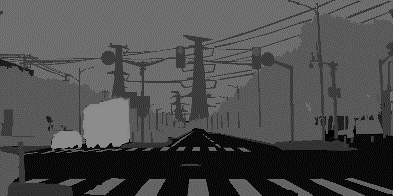}\label{grayscale}}
\caption{Refinement of annotations and processed results.}
\label{fig:refine}
\end{figure}

The example of the initial images, labeled results, and gray-scale maps are shown in Figure \ref{fig:refine}.

\subsubsection{Effect of pre-training}

This paper first trained the PSPNet backbone to make it learn the features of road scenarios. The pre-trained model was built in three steps: 

Step1: Verifying the feasibility of the pre-trained model. This study first replaced the class parameters with the classes of the complex road scenarios in the PSPNet network, Then the network was trained with VOC 2007 dataset (17989 images) to verify the perception accessibility.

\begin{figure}[H]
\centering
\subfloat[Loss value of training set and validation set]{\includegraphics[width=0.4\textwidth]{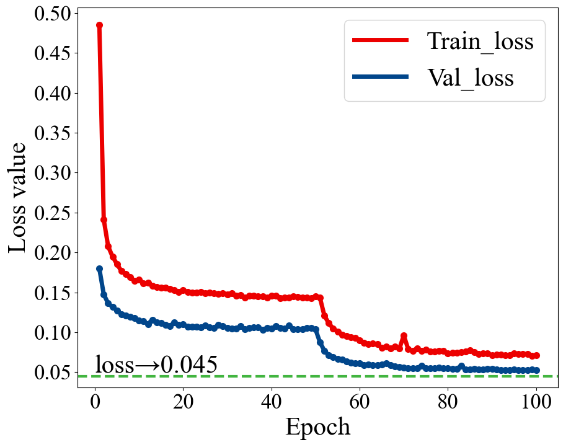}\label{loss}}
\hfil
\subfloat[The ranking of the $mIoU$ of the pre-trained model ]{\includegraphics[width=0.49\textwidth]{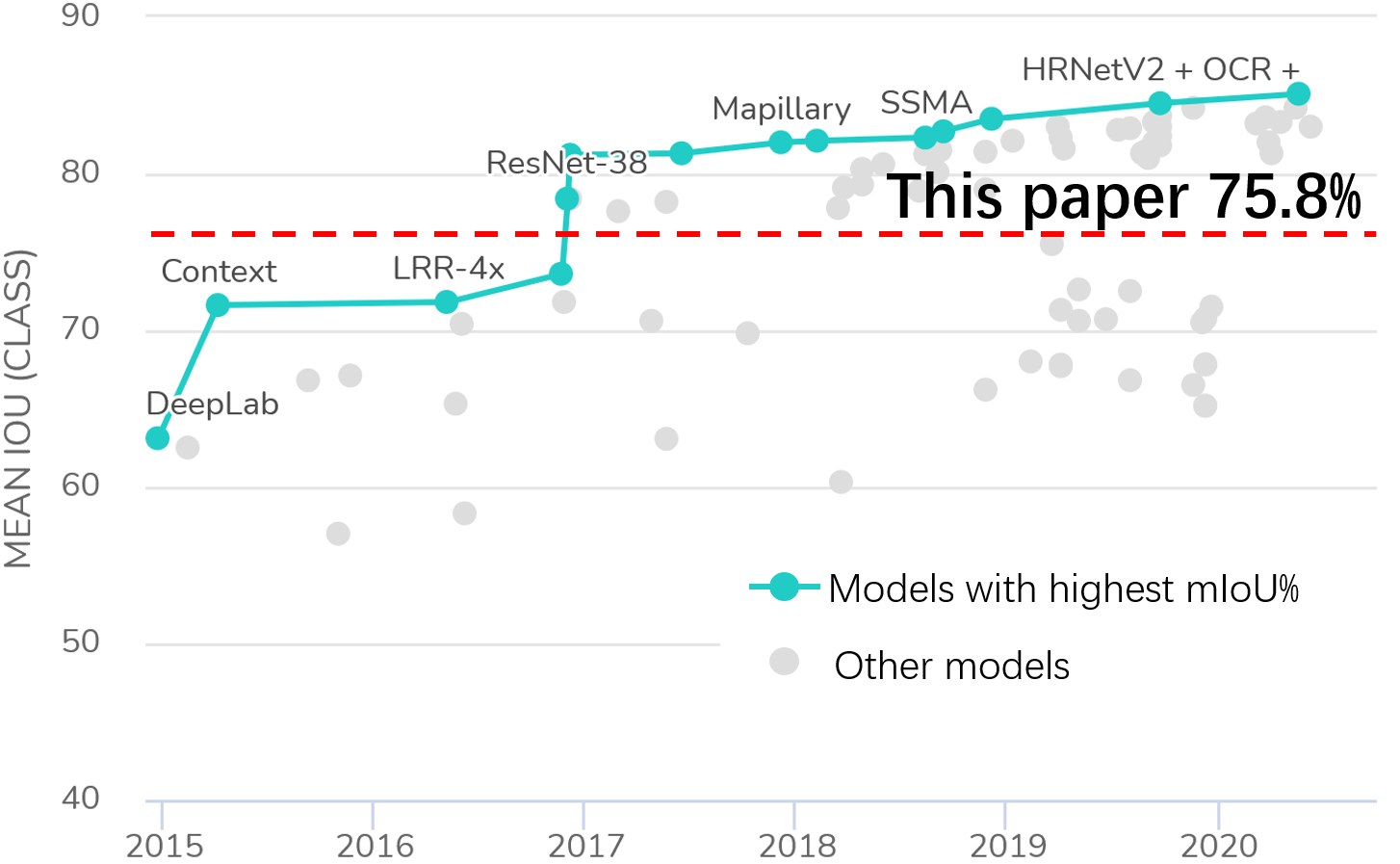}\label{comparison}}

\caption{The feasibility validation of the initial semantic segmentation model.}
\label{fig:initial-results}
\end{figure}

Step2: Coarsely training. Because scenes of VOC 2007 dataset are simple, the pre-trained model cannot adapt to complex road scenarios and detect small size traffic elements well. So the open-source road dataset Cityscapes-Coarse (19998 images) was selected to extract convolutional kernel features. The dataset was randomly divided into the training set and validation set. The ratio of the two sets was 9:1. The loss value results are shown in Figure \ref{loss}. The model converged quickly during the iteration.

Step3: Refine training. After coarsely training, traffic elements could be correctly classified. However, part of the edge information was missed and there were still misclassification and perception failure. So Cityscapes-Fine dataset (5000 images) was applied to improve the extraction of boundary features.

For the validation set of Cityscapes, this study calculated the average mIoU of 22 classes. The segmentation accuracy reached 75.8\%, as shown in Figure \ref{comparison}. It is in the upper level of the current open-source segmentation algorithms. So the feasibility of the pre-trained model was verified.

\begin{figure}[htbp]
\centering
\subfloat[The overall object is fragmented]{\includegraphics[width=0.3\textwidth]{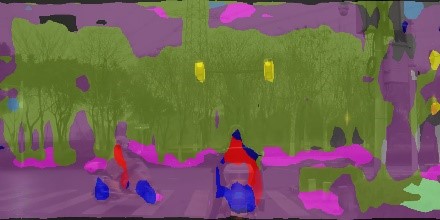}\label{fail-1}}
\hfil
\subfloat[Blurring the boundary of classes ]{\includegraphics[width=0.3\textwidth]{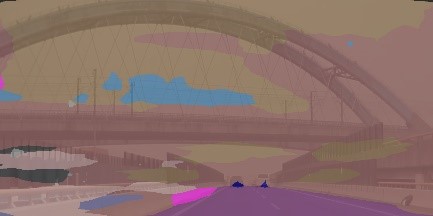}\label{fail-2}}
\hfil
\subfloat[The sky is classified as ground]{\includegraphics[width=0.3\textwidth]{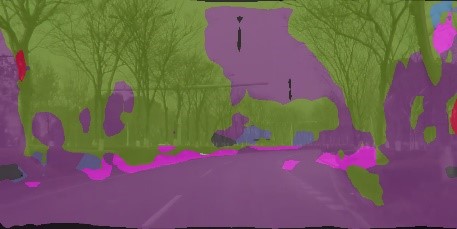}\label{fail-3}}\\
\subfloat[Body reflective perception failure]{\includegraphics[width=0.3\textwidth]{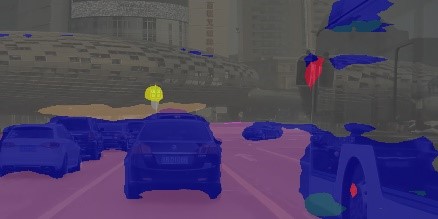}\label{fail-4}}
\hfil
\subfloat[Traffic signals perception failure]{\includegraphics[width=0.3\textwidth]{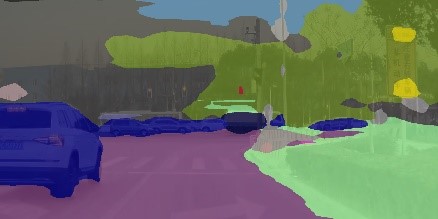}\label{fail-5}}
\hfil
\subfloat[Conflict vehicles' misclassification]{\includegraphics[width=0.3\textwidth]{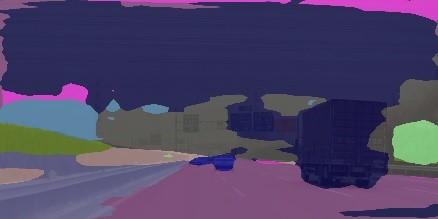}\label{fail-6}}\\

\caption{Perception failures of the pre-trained model in complex road and traffic scenarios.}
\label{fig:fail}
\end{figure}

Furthermore, this study applied the pre-trained model to detect the complex scenario dataset directly. Results suggested that the mean accuracy of segmentation was only 33.25\%. False or missed detection problem were exposed. Figure \ref{fig:fail} shows the examples of perception failure. 

Therefore, it's essential to enhance the segmentation accuracy and perception efficiency for critical traffic elements. This paper improved the segmentation network based on transfer learning. 

\subsubsection{Improved segmentation model based on transfer learning}
Since training the segmentation model with a blank segmentation model required a large number of samples and training time, this paper applied the transfer learning method to improve the accuracy and efficiency of the pre-trained model.

\begin{figure}[H]
\centering
\includegraphics[width= 0.85 \textwidth]{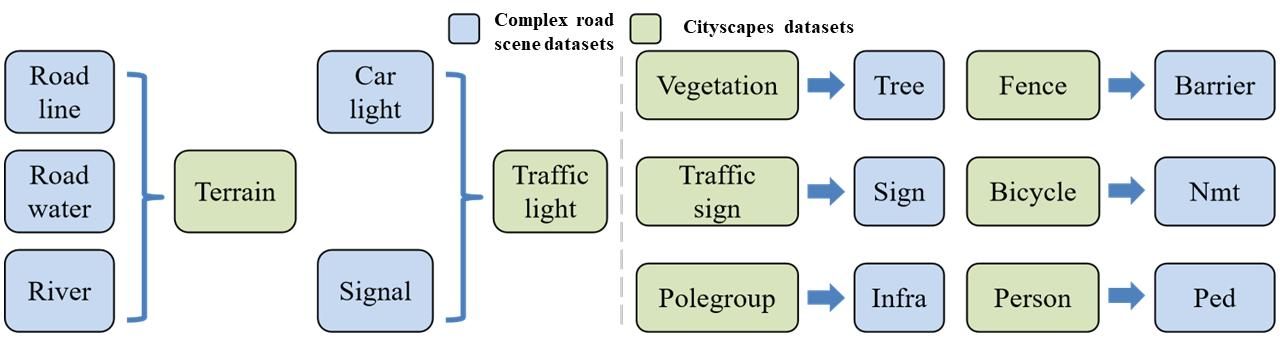}
\caption{Data migration from Cityscapes to the complex road and traffic scenarios.}
\label{fig:trans-1}
\end{figure}

Based on the comparison of two datasets, two reasons caused the poor prediction performance of the pre-trained model. On the one hand, the classification criteria between Cityscapes and the complex scenario datasets are different. So it's difficult for the pre-trained model to perceive the unique classes of the complex scenario dataset. On the other hand, the constitution of the dataset in this paper was much more complex than those in Cityscapes. It led that the pre-trained model was under-fitting for the complex scenarios\citep{DBLP:journals/corr/abs-2112-09298}. To solve the above problems, this paper migrated category data between the two datasets, including the ``many-to-one" and ``one-to-one" mapping relationships, as shown in Figure \ref{fig:trans-1}.

\begin{figure}[htbp]
\centering
\includegraphics[width= 0.88 \textwidth]{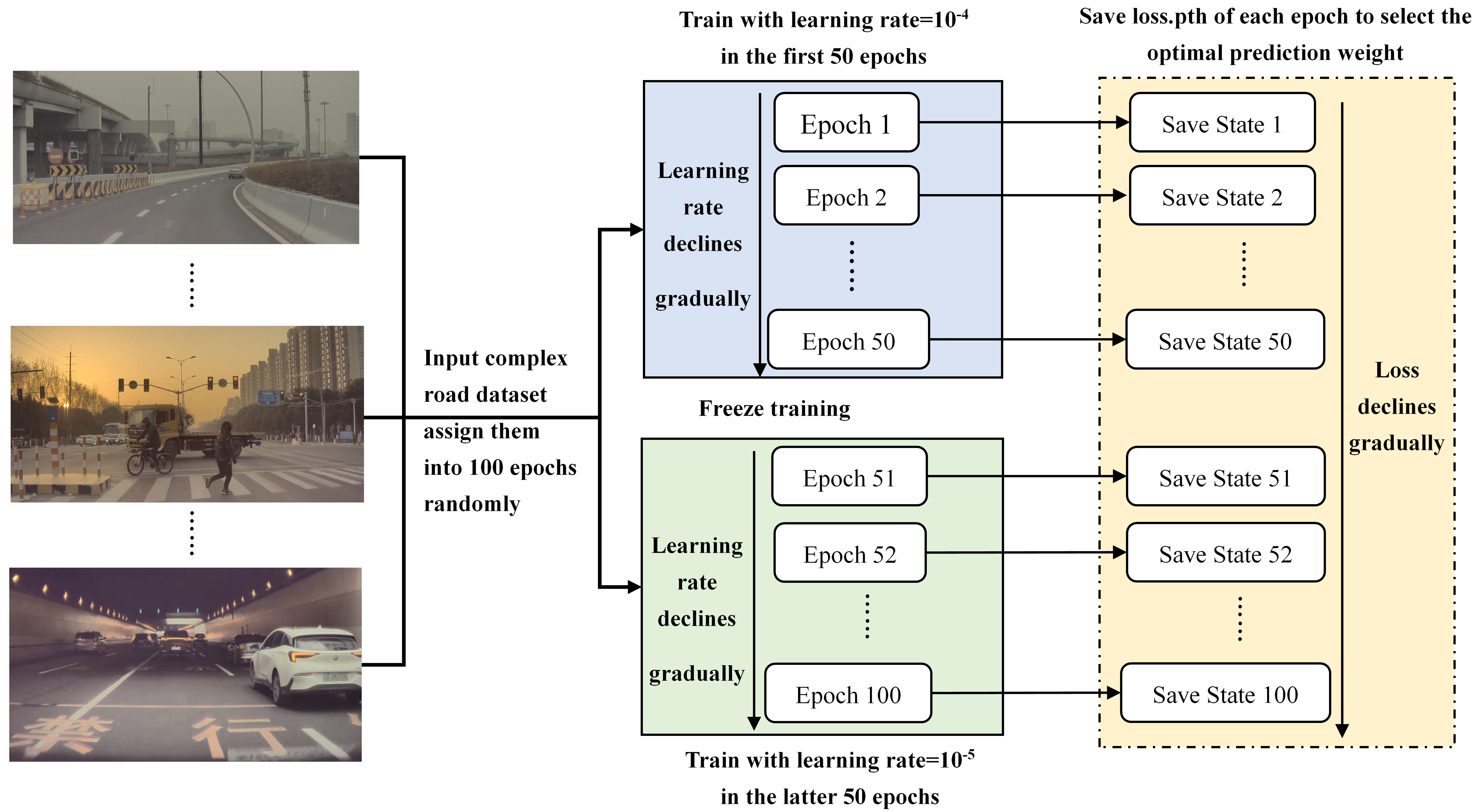}
\caption{The illustration of the freezing and unfreezing processes.}
\label{fig:frozen}
\end{figure}

After data migration between two datasets, this paper also adjusted the hyper-parameters of the pre-trained model, which had been analyzed in Section 3.2.3.
Besides, this paper utilized the freezing training method to enhance the efficiency and robustness of the segmentation model for complex road and traffic scenarios. The features extracted from the neural network were generic. Combined with the data migration, this study froze the completed epochs. It could accelerate the training efficiency and prevent the weights from being corrupted. Figure \ref{fig:frozen} shows the freezing and unfreezing processes.

The 100 epochs were trained once for freezing and unfreezing training. The initial learning rate was $10^{-4}$ for the first 50 epochs while lr of the second 50 epochs was $10^{-5}$. The learning rate would be smaller as the loss function value got closer to the fitness, which ensured the convergence of the model. The state obtained after each epoch was saved separately, and it was used to calculate the prediction weight.

\begin{figure}[H]
\centering
\includegraphics[width= 0.89 \textwidth]{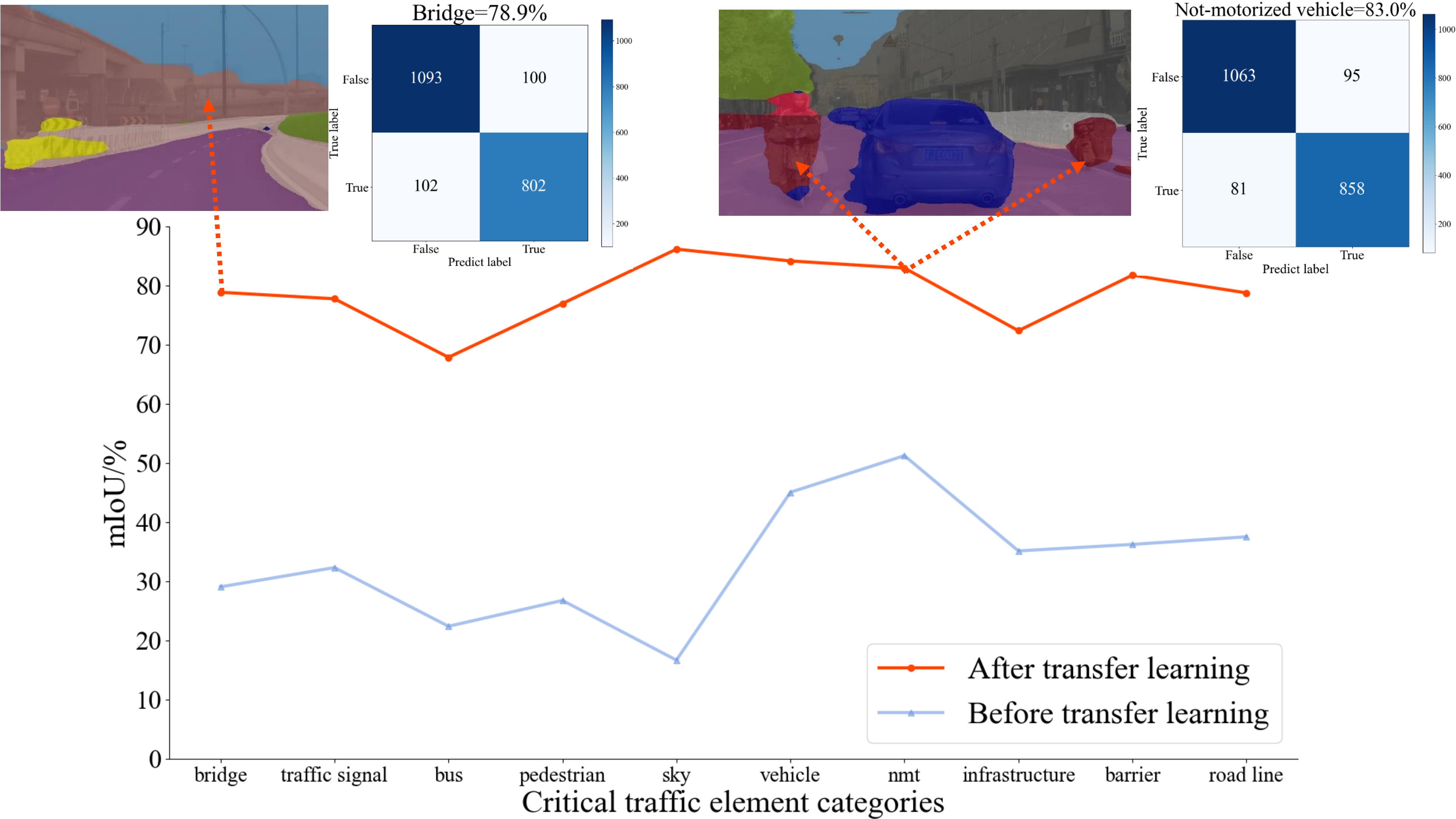}
\caption{Optimization results after transfer learning.}
\label{fig:opti}
\end{figure}

Then, this paper utilized 2097 images to evaluate the effect of model improvement. The $mIoU$ comparison of critical traffic elements in the complex driving environment is shown in Figure \ref{fig:opti}. 

Compared with the prediction results of the pre-trained model, the perception accuracy of each category was improved significantly, especially for the critical traffic elements such as complex road infrastructures, non-motorized vehicles, pedestrians, etc. The average $mIoU$ of the above classes reached 78.80\%, improving by about 45.55\% compared with the $mIoU$ before transfer learning. The result indicated that transfer learning is effective in improving the accuracy of the segmentation model. The improved segmentation model could adapt to complex road traffic scenes.

Besides, the training time decreased from 2.5 hours each epoch (with the initial PSPNet network) to 13 minutes each epoch (with the improved model). It suggested that the improved network could adapt to complex road and traffic scenarios more efficiently.

\subsection{Textual explanation for complex road and traffic scenarios}

Based on the accurate semantic segmentation results, this paper proposed a comprehensive model to obtain the textual information of complex road and traffic scenarios.

\subsubsection{Textual explanation for traffic elements}

Based on the segmentation result, this paper extracted the scenario class to explain what situation AVs are located in. Then the location and state of traffic elements could be analyzed. The model was constructed by the following steps:

\paragraph{Step1: Segmentation data preprocessing}~{}

\begin{figure}[H]
\centering
\subfloat[Semantic segmentation result]{\includegraphics[width=0.45\textwidth]{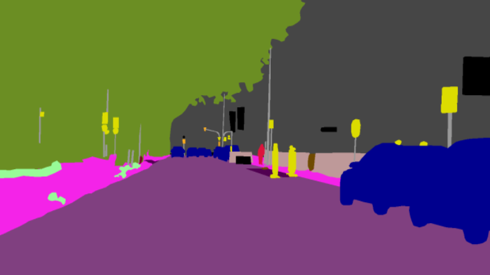}\label{city}}
\hfil
\subfloat[Numerical transforming result]{\includegraphics[width=0.4\textwidth]{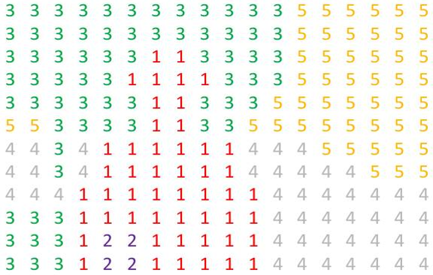}\label{semantic}}

\caption{The preprocessing of segmentation results.}
\label{fig:data}
\end{figure}

First, this paper extracted images of the self-built dataset at a frequency of 1 sheet/10 frames; second, images were input into the improved segmentation model to assign the class labels; then, the segmentation results were converted into the numerical value of class ID. An example of segmentation result preprocessing is shown in Figure \ref{fig:data}.

\paragraph{Step2: Textual information extraction}~{}

\begin{figure}[H]
\centering
\subfloat[The initial image of a complex intersection]{\includegraphics[width=0.5\textwidth]{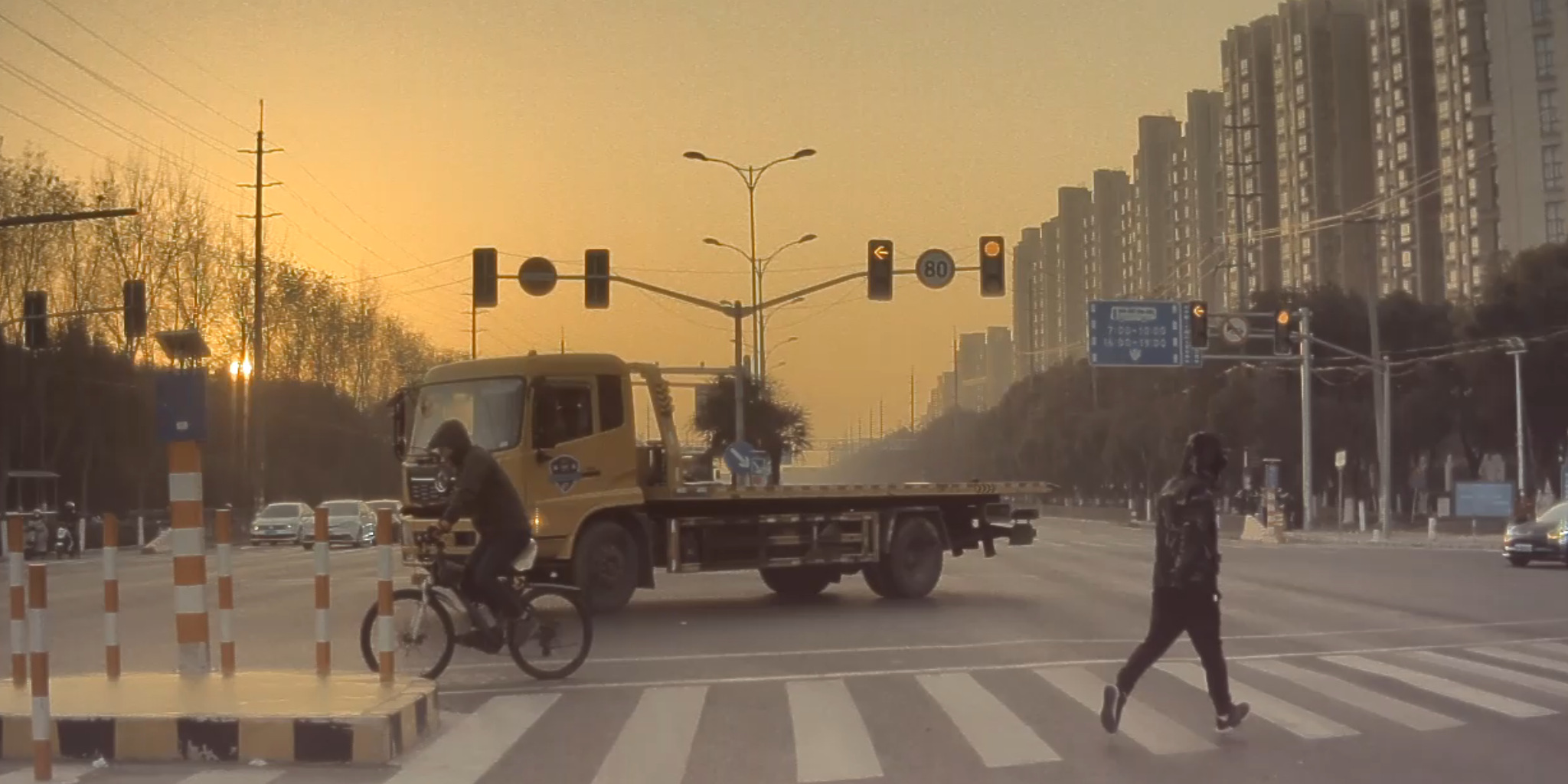}\label{img}}
\hfil
\subfloat[Element extraction result]{\includegraphics[width=0.5\textwidth]{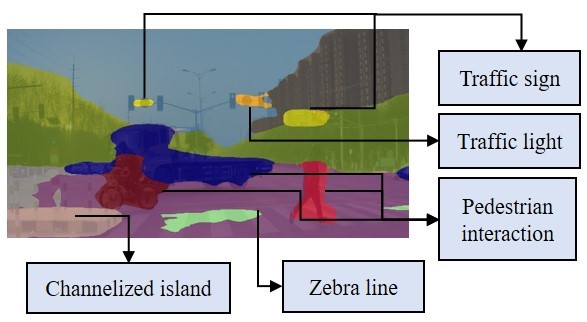}\label{extract}}
\caption{Semantic segmentation feature variables and the example of traffic element extraction.}
\label{fig:extract1}
\end{figure}

Based on the numerical value of segmentation, the visual perception system could obtain class information in a scenario. Figure \ref{fig:extract1} shows an intersection scenario example.

Based on the RGB value of the initial image, the model could return "The traffic light is red, please slow down and stop." The zebra line reflected "There is an intersection, please be careful." And the lateral moving traffic flow meant "There is an interaction in the vertical direction, please keep a safe distance".

\paragraph{Step3: Scenario classification}~{}

Due to the involvement of pedestrian and non-motorized vehicles, this study added relation complexity into consideration. So this paper constructed a scenario classification model based on the preprocessed segmentation data. 

\begin{figure}[H]
\centering
\subfloat[Free driving]{\includegraphics[width=0.23\textwidth]{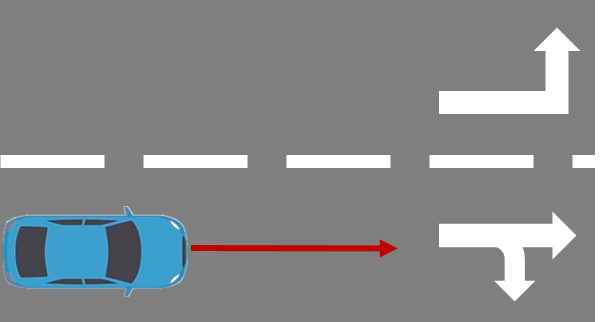}\label{s1}}
\hfil
\subfloat[Following]{\includegraphics[width=0.23\textwidth]{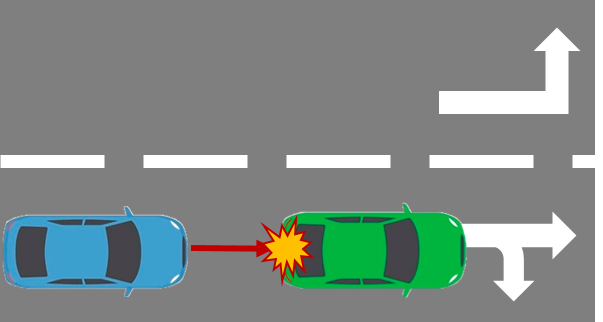}\label{s2}}
\hfil
\subfloat[Cutting in]{\includegraphics[width=0.23\textwidth]{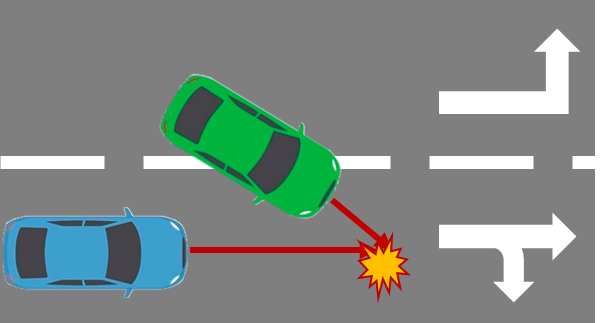}\label{s3}}
\hfil
\subfloat[Emergency Avoidance]{\includegraphics[width=0.22\textwidth]{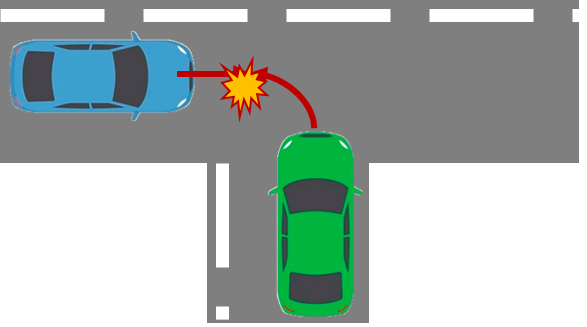}\label{s4}}
\caption{Definition and example of scenario labels.}
\label{fig:label}
\end{figure}

According to traffic elements location information, this paper divided traffic scenarios into four types, as shown in Figure \ref{fig:label}.

\begin{itemize}
\item Free driving: there are no objects in the front direction.

\item Car following: vehicles are driving in front of the subject vehicle in the same lane.

\item Cut-in: vehicles are riding the lane line and driving at a certain angle.

\item Emergency avoidance: vehicles or pedestrians are crossing the road in a vertical direction.
\end{itemize}

Then this paper established the mapping correspondence from the independent variable (traffic elements) to the dependent variable (scenario labels) by the supervised learning method. This study determined feature variables to judge the characteristics of four scenarios. Due to the different distribution features of these elements, this paper established three feature variables, including:

\begin{itemize}

\item Binary variable, i.e., whether an element is present in the scenario, present=1, absent=0. It's mainly for those elements that only exist in the specific scenario.

\item Pixel sum, i.e., the clustered area of pixel points in the same category, e.g., traffic signal has the specific sizes and colors, so the pixel sum is also fixed generally. It's mainly for fix-size elements.

\item Clustering center of gravity coordinates (X, Y), e.g., the gravity center of the zebra line is always in the 1/3 bottom of an image while the gravity center of the sky is always in the 1/3 top of the image. It's mainly for elements that have specific location distribution.

\end{itemize}

\begin{figure}[H]
\centering
\includegraphics[width= 1 \textwidth]{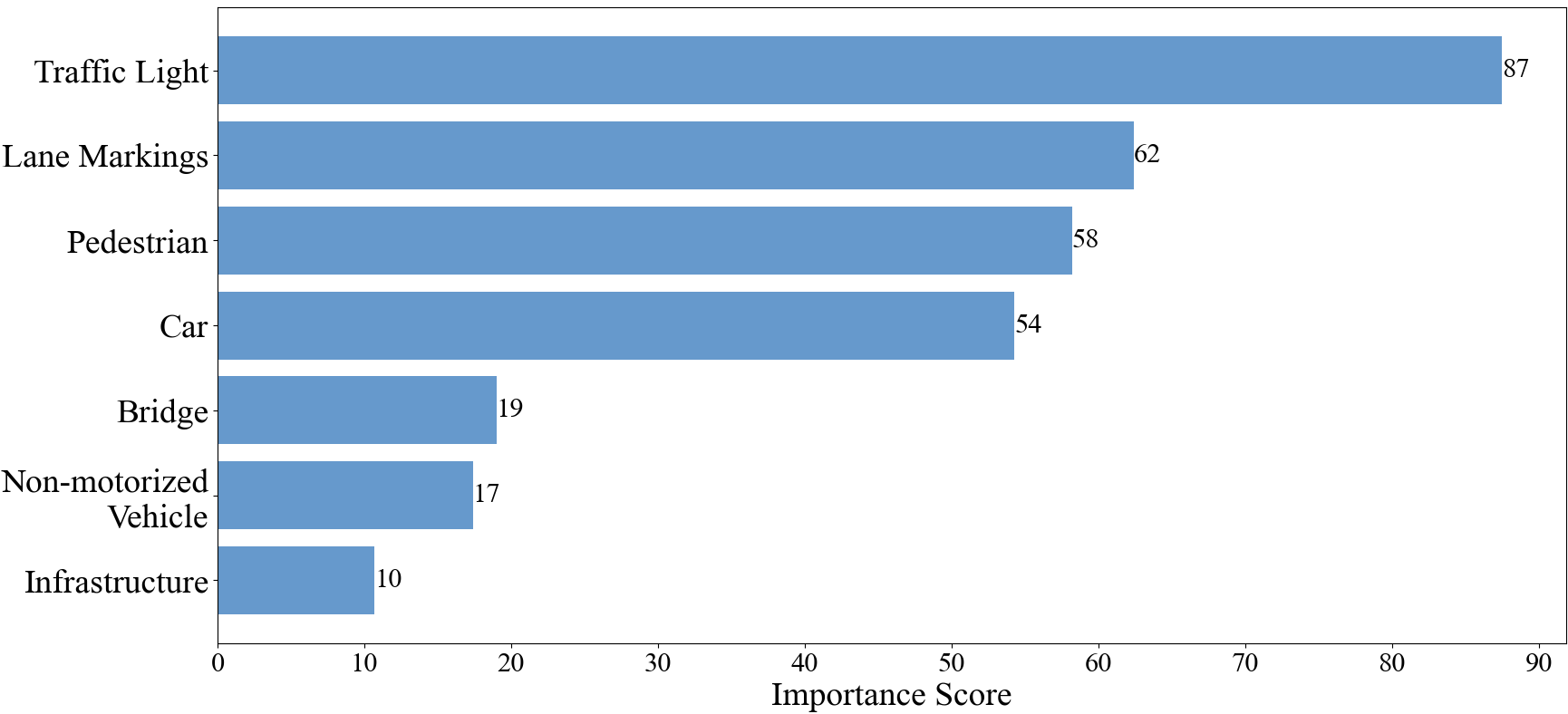}
\caption{The importance of independent variables in the complex road and traffic scenarios.}
\label{fig:importance}
\end{figure}

According to the characteristics of traffic elements, this paper manually associated each image with a scenario label from four types of scenarios. 1500 frames were extracted from the self-built dataset. Through the improved semantic segmentation algorithm, the segmented images were further transferred to numerical values. According to the value of each pixel, this study calculated the number of traffic elements (n=22) and assigned images with scenario labels. Then this study evaluated the importance of the independent variables. Figure \ref{fig:importance} shows the rank result. 

Based on the recursive feature elimination (RFE) and k-fold cross-validation, this study extracts the optimal classification group: car, traffic signals, pedestrian, and lane markings. Applying the group to classify the traffic scenarios, this paper obtained the type of each image. And according to the road infrastructures detection, the model could identify the road scenarios rapidly, including cross, ground, fly-over, ramp, tunnel, and expressway.

\begin{figure}[H]
\centering
\subfloat[Cross-validation confusion matrix of road scenario]{\includegraphics[height=0.4\textwidth]{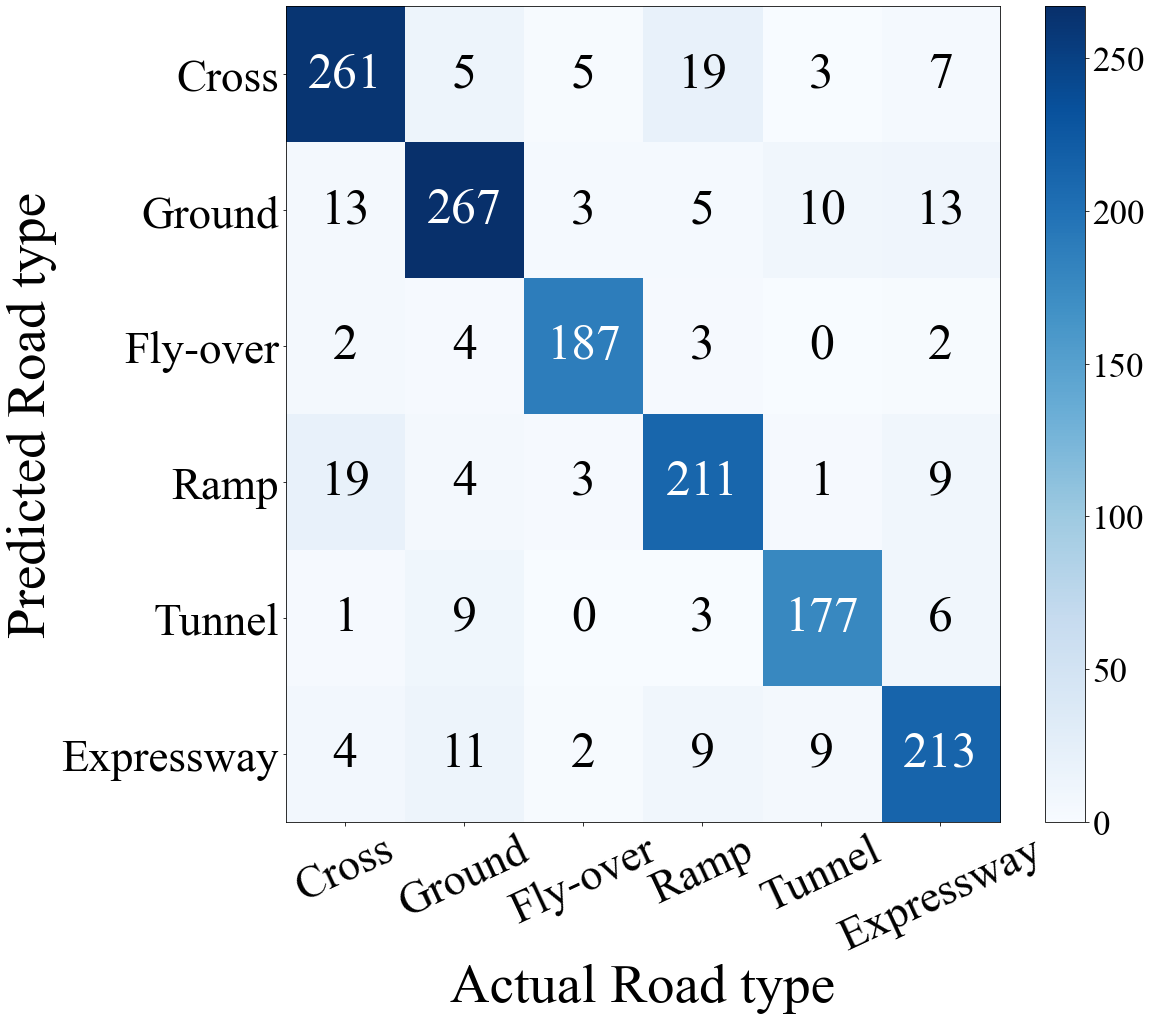}\label{road}}
\hfil
\subfloat[Cross-validation confusion matrix of traffic scenarios]{\includegraphics[height=0.4\textwidth]{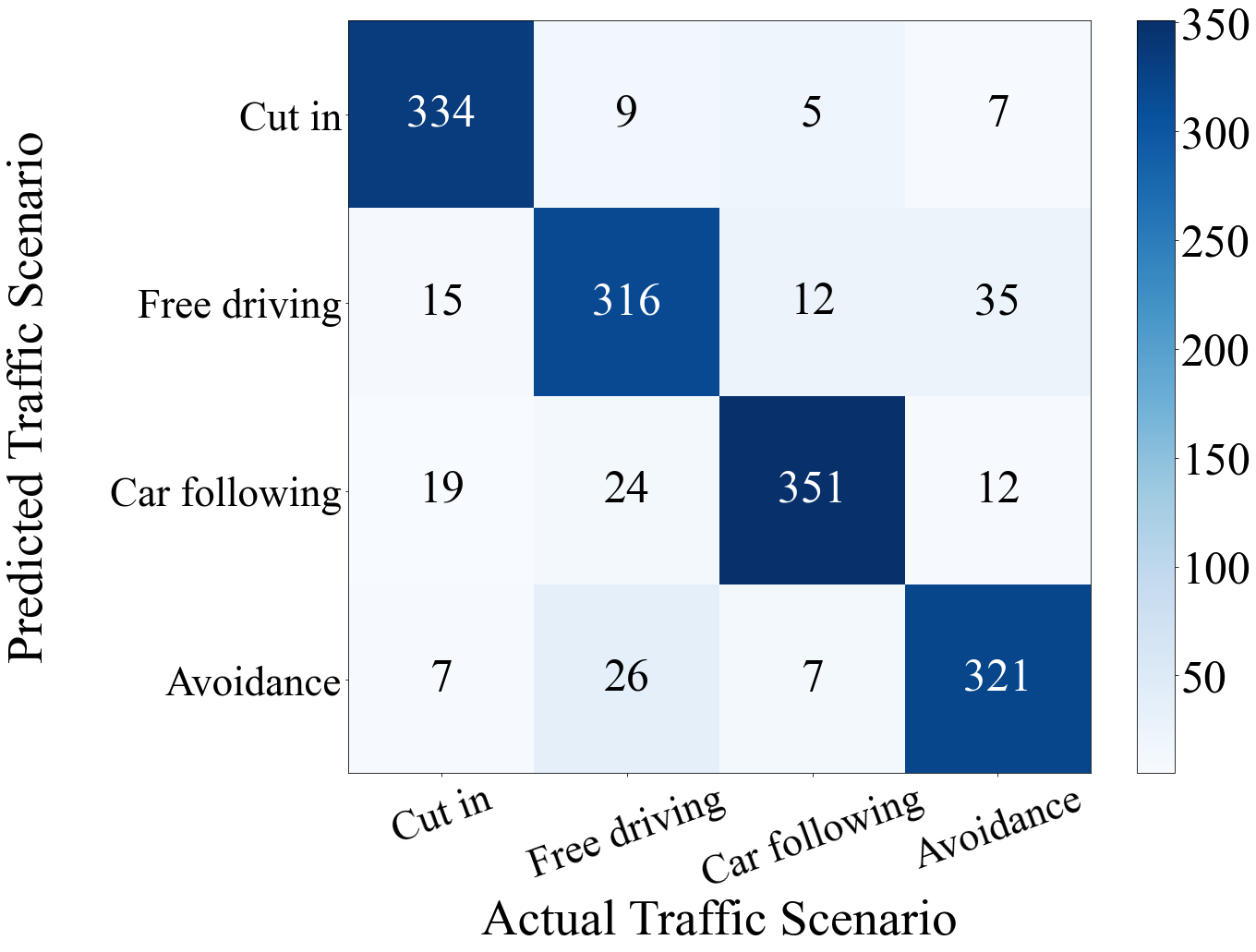}\label{traffic}}

\caption{Cross-validation results of the scenario classification model.}
\label{fig:class}
\end{figure}

To verify the reliability of the model, this study completed the cross-validation of the 1500 images. The results are shown in Figure \ref{fig:class}, with F1-score reaching 88.04\% for road types and F1-score reaching 88.14\% for traffic scenarios. The result indicates that the classification model had high accuracy and provided a reliable relation complexity.

\subsubsection{Estimation of conflict object's motion}

\begin{figure}[H]
\centering
\includegraphics[width= 0.96 \textwidth]{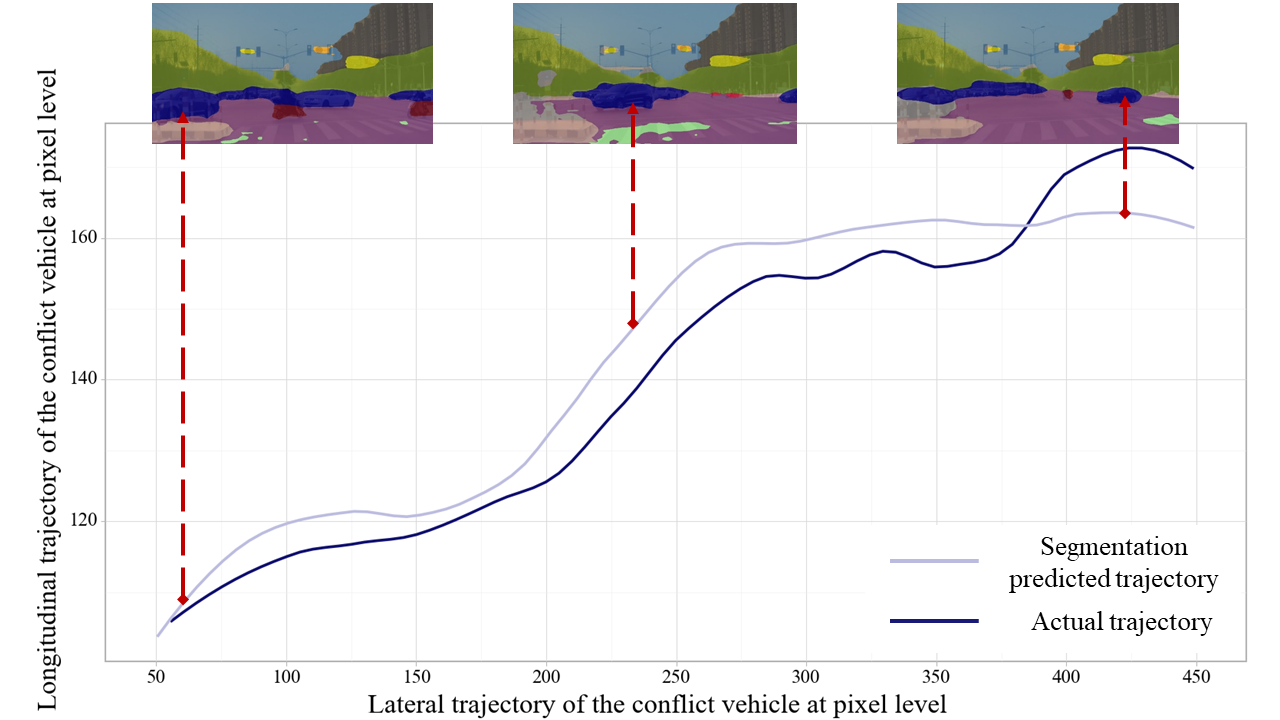}
\caption{Conflict vehicle's motion detection through temporal difference technique.}
\label{fig:traj2}
\end{figure}

Apart from the textual information of traffic elements, estimation of conflict object's motion can evaluate safety level. It offers dynamic clues for subsequent decision-making and  control for AVs. It In this section, this study chose an intersection to detect kinematic states of conflict objects. The images at the top of Figure \ref{fig:traj2} show the scenarios after semantic segmentation. Focusing on the vehicle crossing from the lateral direction, this study selected the clustering center to represent the location of the vehicle and depicted the kinematic trajectory at the pixel level (the dark purple curve in Figure \ref{fig:traj2}). Simultaneously, this study utilized LIDAR to detect the accurate trajectory of the conflict vehicle (the light purple curve in Figure \ref{fig:traj2}). 

When the confidence interval was 95\%, results revealed that the predicted trajectory was very close to reality. The RMSE between two trajectories reached 4.18 pixel$^2$, proving the high precision of kinematic state detection. Based on the accurate trajectory and location information of the conflict car, the velocity and acceleration could be detected by Equations \ref{eq:state} and \ref{eq:ttc}.

\subsubsection{Scenario complexity evaluation}

Based on the segmentation result, motion estimation, and scenario classification, this paper evaluated the global scenario complexity. The classification offered the confidence value of the most likely scenario while TTC revealed the risk level of interaction. The above factors constituted the relation complexity. The number of  traffic elements measured the quantity complexity, and $mIoU$ evaluated the variety complexity. Then images of four road sections were assigned with scenario labels based on the classification model. The probability distribution of each traffic scenario in each section was obtained. According to segmentation accuracy and the number of traffic elements($n_{max}$=22), the model calculated the scenario complexity according to Equation \ref{eq:complex}. To test the scenario complexity evaluation model, this paper selected a path from Jiading District to Yangpu District, Shanghai, China. Four road sections were extracted, including expressway, highway, overhead, and ground road. Table \ref{tab:evaluation} shows the complexity evaluation result compared with the experts.

\begin{table}[H]
\centering
\caption{Results of the textual explanation model compared with the experts}
\begin{tabular}{@{}ccccccc@{}}
\toprule
\multirow{2}{*}{\textbf{Segmentation}} &
\multicolumn{2}{c}{\textbf{Scenario Complexity}} & \multicolumn{2}{c}{\textbf{Relation}} & \textbf{Quantity} & \textbf{Variety} \\
& Model & Expert & Classification & TTC &  $\frac{n}{n_{max}}\times100\%$ & $mIoU$ \\
\midrule
\begin{minipage}{0.1\textwidth}
\includegraphics[width=15.36mm, height=8.64mm]{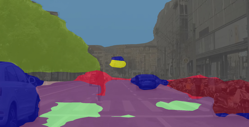}
\end{minipage} & 8.7 & 8.5 & Avoidance 86.5\% & 1.54s &  
86.7\% & 52.1\% \\
& & & & & & \\
\begin{minipage}{0.1\textwidth}
\includegraphics[width=15.36mm, height=8.64mm]{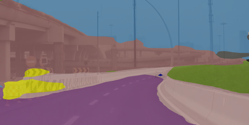}
\end{minipage} & 1.2 & 1.3 & Free driving 80.2\%  & 6.47s & 40.0\% & 77.6\% \\
& & & & & & \\
\begin{minipage}{0.1\textwidth}
\includegraphics[width=15.36mm, height=8.64mm]{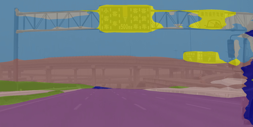}
\end{minipage} & 3.3 & 3.5 & Free driving 89.4\% & 5.31s & 33.3\% & 65.9\% \\
& & & & & & \\
\begin{minipage}{0.1\textwidth}
\includegraphics[width=15.36mm, height=8.64mm]{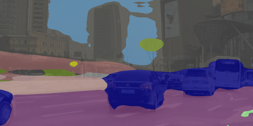}
\end{minipage} & 7.3 & 7.5 & Cut-in 47.4\%  & 1.12s & 75.0\% & 54.5\% \\
\bottomrule
\end{tabular}
\label{tab:evaluation}
\end{table}

To verify the reliability of scenario classification and complexity evaluation results, 18 experts in the field of visual perception were invited to assess the scenarios. The assessment followed the four criteria below.

\begin{itemize}
\item The amount of infrastructure in the scenario and the difficulty of detecting it

\item Classification of driving scenes

\item The type and number of traffic elements to be segmented

\item The driving difficulty from the perspective of human drivers

\end{itemize}

Table \ref{tab:evaluation} shows the comparison between the model and the experts' scores. The MAPE (Mean Absolute Percentage Error) reached 4.84\%. It suggested that the evaluation results of the model were accordant with the trend of the experts' scores. The model could assess the complex driving environment accurately.


\section{Conclusion and Discussion} \label{conclusion}

Focusing on clear and explainable information in the complex driving environment, this study established a textual explanation model. Accurate and efficient semantic segmentation of the driving environment is the foundation of textual explanation. By collecting data in the real-world road environment in Shanghai, China, this study constructed a dataset consisting of complex road and traffic scenarios. With the transfer learning method, this paper made the segmentation network better adapt to complex road infrastructures and mixed traffic flow. The average $mIoU$ of critical traffic elements is enhanced by 45.55\%. By freezing training, the improved model was 11.5 times more efficient than the pre-trained model. Based on the improved segmentation model, the approach could detect the critical traffic elements in the complex environment.

The findings in this paper contribute to explaining what situation AVs are located in and the textual information of traffic elements. It lays a solid foundation for subsequent decision-making and control. The approach can enrich the prior knowledge and judgments for complex traffic situations. Furthermore, scenario complexity evaluation can assist in judging the critical frame of complex environments and constructing a more comprehensive dataset. It benefits future works on more advanced textual explanation frameworks.

Future studies can focus on the refinement of unified textual explanation models. To get a more comprehensive analysis of the traffic situations, the approach should be tested in more complex road and traffic scenarios like more severe weather conditions, traffic congestions, etc. Besides, complexity evaluation can assist in judging the complex road and traffic scenarios and enriching the dataset automatically.

\section*{Acknowledgements}
This study was jointly sponsored by the Chinese National Science Foundation (61803283) and the National Key Research and Development Program of China (2018YFB1600502), and the Shanghai Students' Platform for innovation and entrepreneurship training program(S202110247063).


\newpage

\bibliography{savedrecs}


\end{document}